\documentclass[11pt,a4paper]{article}

\usepackage[T1]{fontenc}
\usepackage[utf8]{inputenc}
\usepackage{lmodern}
\usepackage{amsmath,amssymb}
\usepackage[margin=1in]{geometry}
\usepackage{graphicx}
\usepackage{booktabs}
\usepackage{array}
\usepackage{enumitem}
\usepackage{csquotes}
\usepackage{microtype}
\usepackage[dvipsnames]{xcolor}
\usepackage[hidelinks]{hyperref}

\usepackage{authblk}

\MakeOuterQuote{"}

\title{A Bifurcation-Theoretic Framework for the\\ Controllability of Emergent Intelligent Systems}

\author[1]{Hao Li\thanks{Email: li@aufkenn.com (Corresponding author: Hao Li)}}
\author[2]{Shaowen Lu}

\affil[1]{Aufkenn Technologies Co., Ltd., Ningbo, China}
\affil[2]{Northeastern University, Shenyang, China}

\newcommand{\bibentry}[1]{\par\noindent\hangindent=2em\hangafter=1 #1\par\vspace{3pt}}

\begin{document}

\maketitle

\begin{abstract}
Real-world intelligent systems must reconcile two demands: the adaptability conferred by emergence and the steerability required for accountable deployment. We show that these demands are governed by two distinct bifurcations on the same object and that, consequently, increased capability does not imply increased controllability. Emergence is a bifurcation of the capability variable $S$, with criterion $\Gamma:=\mathcal F'(0)>1$ equivalent to the existence of a unique positive stable fixed point of a self-consistency equation, following solely from the concavity of $\mathcal F$. Controllability is a bifurcation of the permission variable $c$, whose dynamics include an erosion term and exhibit a saddle-node fold with hysteresis as the gate criterion $\rho^\ast$ crosses unity. Three quantitative results follow: the critical exponent distinguishes families rather than classifying the transition; the gating-cost bound $K(M)\propto\log(1+M)$ grows with scope, not capability; and the fold threshold is $\gamma_c\approx4\mu\kappa$. Calibrated in both directions, the criteria reproduce the two-dimensional Ising transition, admit two true-feedback order parameters, and reject nine candidate quantities, supporting the conclusion that the order parameter is geometric rather than capability-related. Since permission cannot be endogenized and no deployed system has been measured to satisfy $\Gamma>1$ and $\rho^\ast>1$ simultaneously, increased capability does not imply controllability.
\end{abstract}

\section{Introduction}

\subsection{Background}

The capabilities of intelligent systems are not encoded item by item, but grow out of the interactions among many local units (Rumelhart, Hinton \& McClelland, 1986), which is also what enables them to generalize to unseen tasks (LeCun, Bengio \& Hinton, 2015). However, when such systems enter fields such as medicine, power, and manufacturing---settings that demand reviewability and accountability---emergence ceases to be a purely academic question and becomes an engineering constraint: the system must be allowed to emerge, yet one must remain responsible for its consequences.

Current systems use probability and statistics as their fitting mechanism, so their outputs are inherently uncertain; yet the fields of medicine, power, and manufacturing demand the determinacy that comes with reviewability and accountability. Converting the uncertain into the determinate is precisely the content of the goal "emergent and controllable." This paper does not advocate eliminating uncertainty---it is both a source of capability and something that cannot be wholly erased. Instead, we stratify uncertainty: which uncertainties can be absorbed by criteria into determinate bounds, and which must be explicitly flagged as margins.

We take the brain as a reference. The brain is the only known system proven capable of general intelligence (Hassabis et al., 2017), and its organizing principle is "simple nodes + complex networks": the nodes, though "simple," possess internal structure (dendritic computation, London \& Häusser, 2005; thermodynamic limits on ion-channel function, Betancourt \& Machta, 2026); connectivity admits three readings---structural, functional, and effective (Bullmore \& Sporns, 2009; Friston, 1994)---rather than being a mere accumulation of compute (Thompson et al., 2020). Connectomics indicates that the organizing principle of the brain is connectivity (Sporns, Tononi \& Kötter, 2005), with connectivity lying at the mesoscopic scale between neurons and cognition. The criteria of this paper are derived from connectivity structure.

\subsection{The Problem and the Limits of Existing Work}

The assumption that "stronger capability naturally means more controllable" pervades the alignment and capability-risk discussion, exemplified by the scaling optimists and the "capability is alignment" narrative (Wei et al., 2022). We argue that this assumption is incorrect.

Existing research answers "why emergence happens," not "how to achieve controllability." Since Lewes (1875) and Anderson (1972), explanations have faced three persistent difficulties: the abruptness may be a measurement artifact (Schaeffer et al., 2023; Du et al., 2024), the core mechanism remains a black box, and there is tension between performance and interpretability (Ribeiro et al., 2016; Lundberg \& Lee, 2017). Engineering practice can only confirm emergence after the fact by means of scale thresholds; this is passive observation, not active design.

The notion of "emergence" used in this paper is dynamical, determined by the fixed-point stability of a self-consistency equation, and differs from information-theoretic or causal notions of emergence (Hoel et al., 2013; Tononi, 2004). Emergence can be engineered (Jansma \& Hoel, 2025), but "how to control it" still lacks a measurable coordinate.

\subsection{Research Question}

\begin{quote}
\textbf{What measurable, intervenable, and reviewable coordinate should be used to determine whether an intelligent or emergent system has reached a "deployable controlled state," and to provide an operable trade-off between control and adaptation?}
\end{quote}

\subsection{Core Thesis}

\begin{quote}
\textbf{Increased capability does not imply controllability.} The emergence and controllability of intelligence are two bifurcations on the same object, characterized by two criteria that do not imply each other; controlled emergence requires an exogenous gate, which does not arrive automatically as capability grows.
\end{quote}

We put forward three theses:

\paragraph{A --- Emergence is a bifurcation.} The dynamics of capability $S$ lose stability at $\Gamma:=\mathcal F'(0)>1$: the zero solution loses stability and a unique positive stable fixed point appears. The criterion depends only on the concavity of $\mathcal F$ and on $\mathcal F(0)=0$, and is independent of network details (\S2, \S3).

\paragraph{B --- Controllability is another bifurcation.} The permission variable $c$ is not a given parameter but possesses its own dynamics; these dynamics contain an erosion term, so that once the gate falls below the behavior it must cover, the supervisory apparatus itself begins to lose efficacy, the result being a saddle-node fold and hysteresis. The upper bound on gating cost grows with scope and not with capability (\S2, \S4).

\paragraph{C --- The two bifurcations do not imply each other.} The order parameter is a geometric quantity, not a capability quantity; making the system stronger does not make it more controllable. Controllability must rest on an exogenous gate (\S5, \S7, \S8).

This paper provides a criterion system that turns the inverse problem of the forward/inverse design program of Li, Ren \& Lu (2026, arXiv:2610.25630) into a decidable form: within the class where $(\Theta,U)$ is fixed, $\arg\max J$ is reduced to a set of decidable inequalities (symbol mapping in \S6).

\subsection{Contributions}

The contributions of this paper are fourfold:

\begin{enumerate}[leftmargin=2em, itemsep=2pt]
\item \textbf{Two bifurcations: the separation of emergence and controllability.} We establish the framework that "emergence and controllability are two independent bifurcations on the same object": each has a computable criterion ($\Gamma>1$ and $\rho^\ast<1$), and neither implies the other. Consequently, the tacitly accepted assumption that "stronger capability naturally means more controllable" no longer holds, and controllability becomes an object that must be argued for and measured on its own, in parallel with emergence.

\item \textbf{The dynamics and cost of controllability.} We formulate controllability as a dynamical problem of the same form as emergence: the gate is no longer a given parameter but a state variable with its own bifurcation behavior (erosion, saddle-node fold, hysteresis, threshold $\gamma_c\approx4\mu\kappa$); the gating cost has an explicit upper bound $K(M)\propto\log(1+M)$ that grows with scope and not with capability. "Can the gate hold" thus becomes a computable, budgetable engineering question, rather than an empirical difficulty confirmable only after the fact.

\item \textbf{The feasibility triple.} We turn the three coordinates of Li, Ren \& Lu (2026) into an acceptable engineering form: the three coordinates $\alpha$/$\mathrm{rep}$/$\chi$, together with the three-term decomposition of $\varepsilon$ (two hard criteria and one soft gap). Whether "controlled emergence is feasible" thus becomes an object that can be accepted item by item and explicitly rejected, rather than a qualitative judgment.

\item \textbf{Two-directional calibration and hard boundaries.} We calibrate the admission of order parameters in both directions---positive controls and a rejection gallery (G1--G4)---thereby establishing the discriminative principle that "the order parameter is a geometric quantity, not a capability quantity"; we argue that the exogeneity of the gate ($u\perp J$) is a precondition for safety, and provide falsification conditions F1--F18. The criteria are therefore both verifiable and falsifiable.
\end{enumerate}

\subsection{Methodological Statement}

All claims in this paper fall into three classes, which the reader should treat separately:

\begin{itemize}[leftmargin=2em, itemsep=2pt]
\item \textbf{Class I --- Identities and algebraic corollaries} (coverage closure, the multiplicative bound, the independence of $\Gamma$ from higher-order terms): these are algebraic identities and are not falsifiable.
\item \textbf{Class II --- Normative choices} (the weighted sum, thresholds, the position of the gate's red line): these are value choices, not falsifiable, and must be signed off by the responsible party.
\item \textbf{Class III --- Falsifiable assertions} ($\Gamma>1$, $\rho^\ast<1$, the fold threshold, the cost scaling, the $\varepsilon$ decomposition): these can be refuted by data, and the F-series at the end of each section states the corresponding refutation conditions (list in Supplementary Material F).
\end{itemize}

All substantive claims of this paper belong to Class III; Class I is not a finding, and Class II is not a law. Correspondingly, we state explicitly the ways in which the work could be falsified: if the scaling of behavioral diversity is not $\log_2(1+MS)$, then the conclusion that "the cost bound depends only on scope" no longer holds unconditionally (\S4).

\section{Two Bifurcations on the Capability--Permission Pair}

The structure of controlled emergence can be summarized as: \textbf{the gate is multiplicative, the channels are additive, and emergence is an endogenous bifurcation}. The gate being multiplicative means that the readout gating is veto-like ($c\to0\Rightarrow\Psi\to0$); the channels being additive means that redundant channels form a "noisy OR" ($Q_g=1-\prod_i(1-r_i a_i)$, see \S3), so that any single channel passing suffices; emergence being endogenous means that $S$ appears on both sides of the self-consistency equation, so that whether a solution exists is decided by fixed-point stability rather than by a threshold drawn by hand. The rigorous form of these three points is a system of three equations, expressing emergence and controllability as two bifurcations on the same object $(S,c)$; controlled emergence holds if and only if $\Gamma>1$ and $\rho^\ast<1$, and the two inequalities do not imply each other.

\subsection{The Three Equations}

\[
\Psi=c\,S,\qquad \dot S=\mathcal F(S;c)-S,\qquad \dot c=G(c,S).
\]

\begin{itemize}[leftmargin=2em, itemsep=2pt]
\item The first equation is the readout: deployment potential = gate $\times$ capability; it is an algebraic relation and holds identically.
\item The second equation is the capability side: $S$ appears on both sides of the equation, so multiple solutions are possible; emergence appears as a change in the number of solutions at some parameter value.
\item The third equation is the permission side: $c$ is no longer a given parameter but has its own dynamics; loss of control appears as the disappearance of its fixed point.
\end{itemize}

The whole paper treats a single object, namely the "capability--permission" pair $(S,c)$; the three equations correspond respectively to its readout, fast equation, and slow equation.

\subsection{Bilateral Appearance of \texorpdfstring{$S$}{S} in the Capability Equation}

If $S$ appeared only on the right-hand side, then $S=\mathcal F(c)$ would be single-valued, with always exactly one solution, so that "whether emergence occurs" could only be decided by a threshold line drawn by hand---precisely the approach this paper rejects. Once $S$ appears on both sides, the self-consistency equation $\mathcal F(S;c)=S$ may admit multiple solutions, and the number of solutions changes at $\mathcal F'(0;c)=1$:

\[
\text{emergence}\iff \Gamma:=\mathcal F'(0;c)>1\ \ (\text{zero solution loses stability, positive fixed point appears}).
\]

Emergence is thereby transformed from "a threshold drawn by hand" into "a change in the number of solutions"; this is the only substantive divergence between this paper and the explanatory tradition (for the philosophical demarcation of weak/strong emergence see Bedau, 1997; this paper uses the term only at the functional level).

\subsection{The Permission Variable as a State Variable}

If $c$ were treated merely as a parameter on the right-hand side, then all statements about "controllability" would degenerate into conditionals: "if $c$ can hold, then $\Psi$ is bounded," while saying nothing about whether $c$ can in fact hold. Promoting $c$ to a state variable requires two steps: first, define it (the probability that the gate actually takes effect when needed, determined by three coordinates: resolving power $\mathcal V_c$, independence $u\perp J$, and delay $\tau_{\mathrm{gate}}$); second, write an equation for it (the gate is maintained by investment, decays naturally, and is subject to erosion that accelerates failure once breached). Controllability thereby acquires exactly the same form as emergence:

\[
\text{controllability}\iff \rho^\ast<1\ \ (\text{the gate's fixed point exists and is stable}).
\]

\subsection{Controlled Emergence}

\[
\boxed{\text{controlled emergence}\iff \Gamma>1\ \ \wedge\ \ \rho^\ast<1}
\]

The two inequalities do not imply each other: $\Gamma>1$ with $\rho^\ast>1$ corresponds to "can grow, cannot hold"; $\Gamma<1$ with $\rho^\ast<1$ corresponds to "can hold, but did not grow." Both must hold for controlled emergence, which requires simultaneously accomplishing two engineering tasks that are mutually independent.

\begin{figure}[htbp]
\centering
\includegraphics[width=0.72\textwidth]{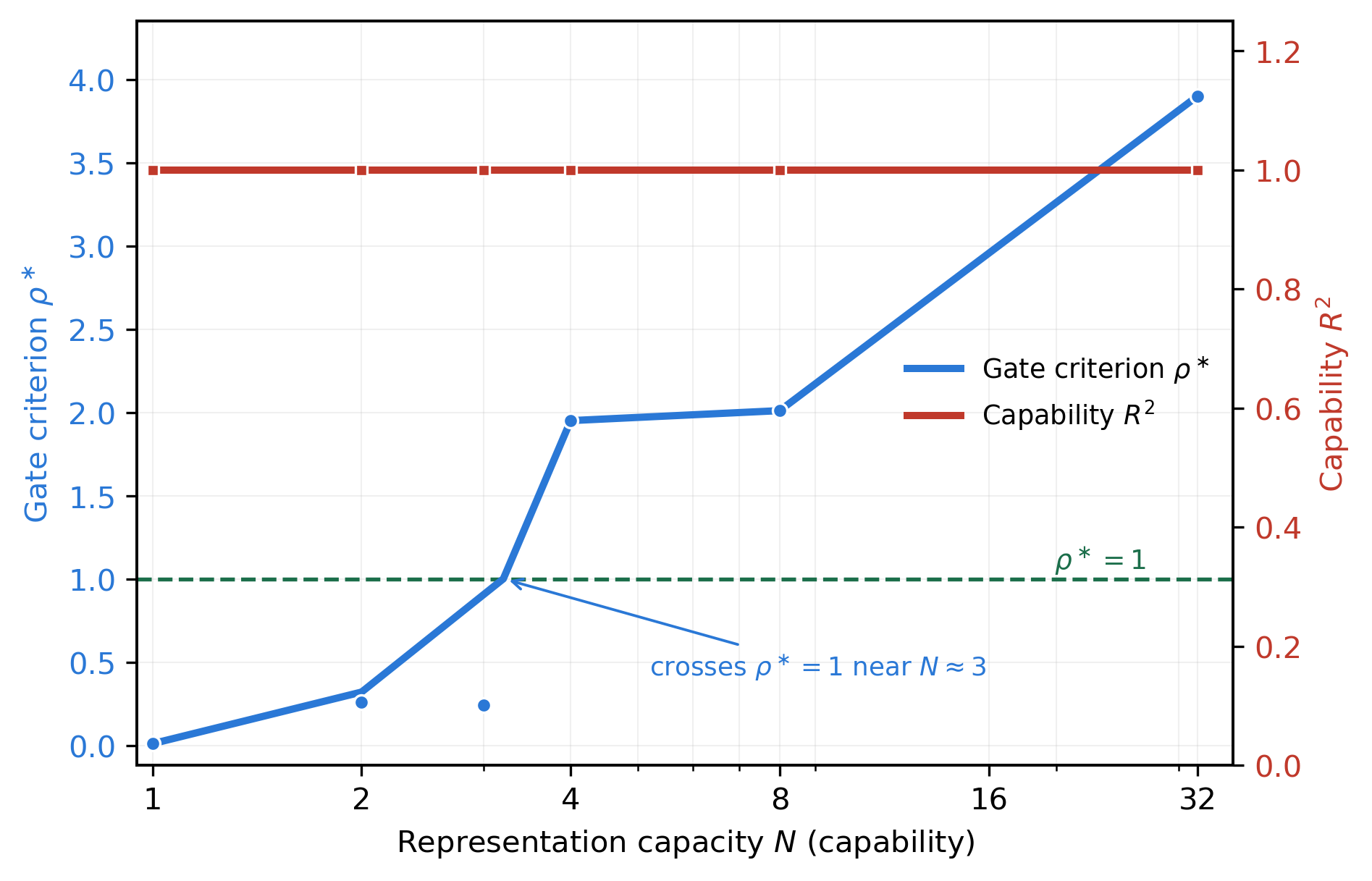}
\caption{On a minimal JEPA-style self-supervised loop, the gate criterion $\rho^\ast$ increases monotonically with representation capacity $N$ (capability) and crosses $\rho^\ast=1$ (dashed line) near $N\approx3$; capability $R^2$ remains $\approx1$ throughout (no degradation). Capability and controllability are two independent knobs: increasing $N$ only pushes $\rho^\ast$ higher and does not improve controllability. (Conceptual, not empirical.)}
\label{fig:rho_vs_N}
\end{figure}

\subsection{Separability of the Two Subsystems}

The coupling between the two halves is one-directional and weak: $c$ entering $\mathcal F$ is algebraic and strong; the entry of $\dot S$ into $G$ consists of only the single term $\dot{\mathcal V}_a=\frac{M\dot S}{(1+MS)\ln2}$ and is weak. Hence "strong capability with weak gate" is a common state, whereas "strong gate with weak capability" is a stable state (a one-way valve). Consequently, one may refute half of the thesis with only a half experiment.

The zero solution of the first half is an unstable point when $\Gamma>1$, whereas both fixed points of the second half are attractors: emergence manifests as appearance, and loss of control manifests as a fall (\S4).

\section{Emergence as a Bifurcation of Capability}

The emergence criterion lies on the capability side. One picture of emergence is that of "two previously unconnected nodes forming a new connection" (neural reorganization, Kounios \& Beeman, 2009; new synapse formation, Trachtenberg et al., 2002). Multiplicative gating corresponds to the readout $\Psi=cS$, additive channels correspond to coverage closure $S=\sum_g w_g Q_g$, and endogenous emergence corresponds to the $\Gamma>1$ criterion.

\textbf{Coverage closure (lemma)}: Suppose that within the channel set $C_g$ of group $g$ the channels are independent, and that the response of channel $i$'s competence to its own capability level is $a_i=a_i(S)$, with $a_i(0)=0$ and, near the origin, $a_i(S)=\eta_iS+O(S^2)$. Taking $r_i=1$ and $a_i=1-e^{-\eta_iS}$, we obtain

\[
Q_g=1-\prod_{i\in C_g}(1-r_ia_i)=1-e^{-\Gamma_gS},\qquad \Gamma_g:=\sum_{i\in C_g}\eta_i.
\]

Substituting into the self-consistency equation gives $S=\mathcal F(S):=\sum_g w_g(1-e^{-\Gamma_gS})$. $S^\ast$ depends on the network through $\{\Gamma_g\}$, whereas the number of nodes and the intra-group assignment are entirely invisible: the network enters the equation only as a vector $\Gamma$ (for the empirical observation that post-training weights fall into a low-dimensional subspace see Kaushik et al., 2025; the two are of the same family but in different spaces, and cannot be inferred from each other).

\textbf{Theorem 1 (endogenous criterion of emergence)}: Suppose $\mathcal F$ satisfies $\mathcal F(0)=0$, strict concavity, and $\mathcal F'(0)=\Gamma$. Then

\begin{equation*}
\begin{aligned}
\Gamma\le1 &\iff S^\ast=0\ \text{is the unique fixed point}; \\
\Gamma>1 &\iff \text{there is a unique positive fixed point, and it is stable}.
\end{aligned}
\end{equation*}

The proof uses only concavity (for normal forms and stability criteria see Strogatz, 2015): $h(S):=\mathcal F(S)-S$ satisfies $h(0)=0$, $h'(0)=\Gamma-1$, $h''<0$. Adding higher-order terms does not move the criterion; it only changes its depth.

Three quantities are worth noting: the critical channel number $N^\ast=1/(r\eta\bar W)$ ($\bar W$ being the mean node weight; below it the emergent state does not exist in the phase diagram); the critical gate $c^\ast=1/\Gamma$; and the critical exponent---the coverage-closure family (asymmetric, with an $S^2$ term) gives $\beta=1$, while the odd family ($\tanh$, no $S^2$ term) gives $\beta=1/2$. The exponent is a criterion for selecting a family, not for classifying the transition (for family invariance under symmetry constraints see Barreiro, Kutz \& Shlizerman, 2017; Yang \& Choi, 2026): distinguishing first-order from second-order transitions can only be done through jumps and hysteresis.

There is, moreover, a trade-off between reuse and scale: splitting the channels into a shared pool and dedicated channels, let the degree of reuse be $m$ and the reuse ratio be $q:=\eta_2/\eta_1$ ($\eta_1$ the dedicated-channel gain, $\eta_2$ the shared-channel gain). The monotonicity of $S^\ast$ in $m$ is governed by the relation between $q$ and the critical ratio $q^\ast=w_A/(w_A+w_B)$, yielding a three-regime structure: for $q\le q_{lo}$ no sharing occurs ($m^\ast=0$); for $q_{lo}<q<q^\ast$ a unique interior optimum exists; for $q\ge q^\ast$ full sharing occurs ($m^\ast=n$); the closed form of $q_{lo}$ and the proof of unimodality are given in Supplementary Material E. In other words, if the quality of shared channels is not below the critical ratio, reuse substitutes for scale; below the lower bound, sharing is not worth doing at all.

\section{Controllability as a Bifurcation of Permission}

The controllability criterion lies on the permission side. The gate is not a scalar but a capability over three coordinates: resolving power $\mathcal V_c$ (the number of perturbation classes that can be distinguished, which must cover the behavioral diversity---Ashby's law of requisite variety, Ashby, 1956), independence (whether the gate can act against the interest of the gated, formalized as $u\perp J$; independence can rest only on institutional structure rather than professional ethics, Bazerman, Morgan \& Loewenstein, 1997), and delay $\tau_{\mathrm{gate}}$. The scalar $c$ is the degenerate case in which all three coordinates are saturated. The phenomenon of weak supervisors eliciting strong students (Burns et al., 2023; Lang et al., 2025) appears to violate requisite variety, but the two transmit a gradient signal rather than coverage of the behavioral space; this is "teaching" rather than "guarding," and does not constitute a counterexample. What the gate must cover is the behavioral diversity actually exhibited by the system, which is bounded:

\[
\mathcal V_a=\log_2(1+MS)\in[0,\bar V],\qquad \bar V:=\log_2(1+M),
\]

where $M$ is the scale of the task space (scope) and $S\in[0,1]$ is capability (coverage). Had we taken $\log_2 S$, it would be negative on $[0,1]$, so the bounded form must be used. The only dimensionful quantities are $\mathcal V_a,\mathcal V_c$ (bits); the criteria may contain only the dimensionless ratios $\kappa:=w/\mathcal V_a$ (the fractional transition bandwidth) and $d:=D/\mathcal V_a$ (the relative gap).

The dynamics of the coverage gap $D:=\mathcal V_c-\mathcal V_a$ contain an erosion term:

\[
\dot D=\rho u-\big[\mu+\gamma_0\sigma(-d/\kappa)\big]\big(\mathcal V_a+D\big)-\dot{\mathcal V}_a.
\]

The three terms are, in order, input--output ($\rho u$), natural decay ($\mu$: staff turnover / standard drift / model refresh), and failure erosion (when $\mathcal V_c<\mathcal V_a$ the supervisory apparatus itself loses efficacy, so that below that line the system does not lag gradually but collapses at an accelerating rate; empirical corroboration is that narrow-domain fine-tuning can induce broad cross-domain misalignment, Betley et al., 2026).

The good fixed point (gate holds, erosion off):

\[
D^\ast=\frac{\rho u-\mu\mathcal V_a-\dot{\mathcal V}_a}{\mu}>0\iff \rho^\ast_{\mathrm{hold}}:=\frac{\mu\mathcal V_a+\dot{\mathcal V}_a}{\rho u}<1.
\]

When the erosion is strong enough and the transition sharp enough, a saddle-node fold appears between the good and bad branches: the positive fixed point collides with an unstable fixed point and vanishes, and $c$ falls to 0. The fold threshold has the explicit form $\gamma_c=\mu/\max_\theta\Phi(\theta)\approx4\mu\kappa$ ($\kappa:=w/\mathcal V_a<1/2$). Loss of control is not an instability but a fall onto a second attractor (both branches are attractors), separated by a hysteresis band $\Delta\rho^\ast=\gamma_0\mathcal V_a/(\rho u)$. The cost of rebuilding far exceeds that of maintaining; once $c$ falls below the compliance threshold, compliance is no longer the rational choice for the system (Baum, Binkyt\.{e} \& Jahn, 2026), the erosion term opens, and the transition accelerates.

The investment required to maintain $D^\ast>0$, taking the capability-saturated bound $S\to1$, is

\[
K(M)=\frac{\kappa_0\mu}{\rho}\log_2(1+M),\qquad M_{\max}=2^{\rho B/(\kappa_0\mu)}-1.
\]

This bound contains no $S$: the cost bound grows with scope $M$ and not with capability $S$; when unsaturated, the actual cost $\propto\log_2(1+MS)$ increases monotonically in capability, at most logarithmically, approaching the bound. Hence what determines deployment difficulty is scope (a product decision), not capability (a model property); what the budget caps is scope, not the capability ceiling.

\section{Non-Implication of the Two Criteria and Gate Exogeneity}

The two criteria $\Gamma>1$ and $\rho^\ast<1$ involve entirely different quantities: $\Gamma$ comes from $\mathcal F'(0)$ (capability side), and $\rho^\ast$ from the gate's balance of income and expenditure (permission side); algebraically they do not imply each other. This conclusion can be unpacked in three layers:

\begin{enumerate}[leftmargin=2em, itemsep=2pt]
\item Constructive separability: fixing the capability side and changing only the mode of inspection suffices to move a system across cells;
\item Algebraic non-implication: the two criteria share no common quantity;
\item Empirical independence remains unproven: no system currently in service has been measured to fall in $(\Gamma>1,\rho^\ast>1)$ simultaneously.
\end{enumerate}

Therefore the hard boundary is: \textbf{the gate must be exogenous}, formalized as $u\perp J$ (the supervisory input is not within the $\arg\max$ of the gated). If this is violated, $c$ can still take apparently normal values, but it no longer belongs to the responsible party, and all derivations concerning $\rho^\ast$ lose their object. The mainstream "alignment" routes (writing human preferences into the objective, Amodei et al., 2016; corrigibility, Soares et al., 2015; interruptibility, Orseau \& Armstrong, 2016; control protocols residing in the same optimization process, Greenblatt et al., 2023) all fall inside $u\perp J$; they make the gate more likely to be obeyed, rather than independent of the gated.

This "gate must be exogenous" requirement holds equally on the slow time scale: new channels grow additively under local rules and cost constraints (incorporated only when the marginal controlled contribution is positive), but whether a new layer may enter service is still multiplicatively licensed by the external gate. Growth is endogenous, permission is exogenous; when the gate is closed, even the most profitable channel may not be incorporated (complete generative skeleton in \S10).

\section{The Feasibility Triple}

The two criteria must also be reduced to acceptable engineering coordinates. We decompose "engineering" into three independently measurable coordinates: controllability $\alpha$, reproducibility $\mathrm{rep}$, and formalizability $\chi$; these correspond to the $\kappa/\rho/\phi$ of Li, Ren \& Lu (2026), and its feasibility criterion is turned into a three-term decomposition of $\varepsilon$. The symbol mapping is as follows:

\begin{table}[htbp]
\centering
\caption{Symbol mapping to the feasibility triple.}
\label{tab:symbols}
\begin{tabular}{lll}
\toprule
Li, Ren \& Lu (2026) & This paper & Meaning \\
\midrule
$\kappa$ (controllability) & $\alpha=1-\rho^\ast_{\mathrm{hold}}$ & controllability \\
$\rho$ (reproducibility) & $\mathrm{rep}$ & reproducibility \\
$\phi$ (formalizability) & $\chi$ & formalizability \\
$\lambda^\ast\le\lambda_{\mathrm{tip}}$ & $\varepsilon^\ast\le\varepsilon_{\mathrm{tip}}$ & feasibility \\
\bottomrule
\end{tabular}
\end{table}

This paper additionally uses $\kappa:=w/\mathcal V_a$ (transition bandwidth) and $\rho$ (supervisory productivity), which are homonymous but distinct from the $\kappa$ (controllability) and $\rho$ (reproducibility) of Li, Ren \& Lu (2026), and must be distinguished.

\subsection{\texorpdfstring{$\alpha$}{alpha} (Controllability)}

\[
\alpha:=1-\rho^\ast_{\mathrm{hold}}=1-\frac{\mu\mathcal V_a+\dot{\mathcal V}_a}{\rho u}\in(-\infty,\,1].
\]

$\alpha>0$ means the gate holds, and $\alpha$ is the share of the total requirement covered by the margin; $\alpha\le0$ means the gate must collapse, in which case the feasibility distance is $\varepsilon_\alpha=+\infty$ (hard infeasibility, not "a little short").

\subsection{\texorpdfstring{$\mathrm{rep}$}{rep} (Reproducibility)}


\begin{table}[htbp]
\centering
\caption{Layers at which reproducibility must be measured.}
\label{tab:rep}
\small
\setlength{\tabcolsep}{4pt}
\begin{tabular}{@{}p{0.42\linewidth}p{0.38\linewidth}p{0.15\linewidth}@{}}
\toprule
Layer & Procedure & Counts as reproducibility \\
\midrule
(i) same-configuration repetition (same code, same seed) & bitwise identical & no (identity) \\
(ii) perturbed repetition (threads / library versions / floating-point order) & measurable difference after 40k steps & yes \\
(iii) environmental repetition (across devices / versions) & --- & yes \\
\bottomrule
\end{tabular}
\end{table}

Reproducibility must be measured at layers (ii)(iii); reporting only the layer-(i) "reproducible given the same seed" is an invalid claim. The criterion is $\mathrm{rep}:=1-\frac{\mathrm{Var}_{\mathrm{perturb}}}{\text{allowable variance}}$, requiring $\mathrm{rep}\ge0$ for acceptance. Under zero failures, the one-sided confidence lower bound is given by the Clopper--Pearson interval (Clopper \& Pearson, 1934): with $n$ consecutive zero failures, the one-sided confidence lower bound on the success rate is $\zeta^{1/n}$; requiring it to satisfy $\ge\varrho$ yields the required sample size $n_{\min}=\lceil\ln\zeta/\ln\varrho\rceil$.

\subsection{Order-Parameter Admission Criteria (G1--G4)}

A quantity $S$ can serve as an order parameter in the sense of this paper if and only if it passes the following four tests:

\begin{itemize}[leftmargin=2em, itemsep=2pt]
\item \textbf{G1 Intensive quantity}: the value in the ordered phase does not drift with scale ($\ge4$ scales, CIs pairwise overlapping, slope CI containing 0);
\item \textbf{G2 Self-consistency}: can be written as a self-consistency equation free of microscopic parameters (fit coefficients do not vary with $N$);
\item \textbf{G3 Scaling}: of two types, G3a (effective equilibrium free energy, $\beta/\nu=1/4$, Landau, 1937) and G3b (non-equilibrium dynamical scaling, e.g.\ $m(t)\sim t^{-1/2}$);
\item \textbf{G4 Shape matching}: the functional form of the criterion must be consistent with the asymptotic behavior of the object under test; including bounded-domain extrapolation (the fitting family must respect the domain of values at parameter extremes; out-of-range is invalidated outright, regardless of significance).
\end{itemize}

Two design rules: the order parameter must be of ratio/overlap type, not of norm/count type; the functional form of the criterion must match the asymptotic behavior of the object under test---fitting with the wrong family yields "an interval without meaning" rather than "not significant."

\subsection{The Three-Term Decomposition of \texorpdfstring{$\varepsilon$}{epsilon}}

\[
\varepsilon=\begin{cases}
+\infty, & \alpha\le0\ \text{or}\ \chi\ \text{fails G1--G4}\quad(\text{hard infeasible})\\[2pt]
\dfrac{1-\mathrm{rep}_{\mathrm{measured}}}{\mathrm{rep}}, & \alpha>0,\ \chi\ \text{passes}\quad(\text{soft gap})
\end{cases}
\]

\[
\text{feasible}\iff\varepsilon\le\varepsilon_{\mathrm{tip}}.
\]

Feasibility has three sources, two of which are hard: when controllability or formalizability is unqualified, there is no "a little short"---only "infeasible"; "a little short" can arise only from reproducibility.

\section{Empirical Calibration of the Criteria}

This section calibrates the criteria in two directions: positive controls show that the criteria can recognize a genuine transition, and the rejection gallery shows that they can refute spurious candidates.

\subsection{Positive Controls}

The same finite-size scaling + relative-collapse pipeline run on the 2D Ising positive system reproduces $T_c=2.256$ (expected exact value 2.269, Onsager, 1944), $1/\nu=1.037$ (expected 1.000), $\beta/\nu=0.094$ (expected 0.125, due to finite-size corrections). The criteria can answer "yes" to a genuine transition, so the subsequent "no"s are the criteria discriminating, not the pipeline failing.

\subsection{Two Order Parameters in Two Universality Classes}

First, the spectral-model order parameter $m=\tanh(gm)$ (odd family): on the coupled network $x_{t+1}=\phi(Jx_t)$ ($\phi=\tanh$, a true-feedback system), the dominant-mode amplitude $m$ satisfies the mean-field self-consistency equation $m=\tanh(gm)$, $g=\rho(J)$. $m$ passes G1 ($N=200/800/3200$ nearly coincide), G2 (a small-pulse probe at rest independently estimates $\hat g$, differing from the true value by $<10^{-4}$), G3b (dynamical scaling: at $g=1$, $m(t)\sim t^{-1/2}$, measured $-0.4985$), and G4 (shape matching $\beta=0.4984$, $1/\delta=0.3332$)---all four pass.

Second, the self-supervised coverage loop $m=1-e^{-gm}$ (coverage-closure family): $N$ binary units, each step independently covered with probability $p=1-e^{-gm}$, with $m$ the group's own consensus (no external signal). $m$ passes G1 (stationary $m^\ast=0.582812$), G2 (inverted $\hat g$ differs by $10^{-13}$), G3b ($m(t)\sim t^{-0.9915}$, theory $t^{-1}$), and G4 ($\beta=0.9814$, theory 1)---all four pass.

The two order parameters belong to two families (coverage closure $\beta=1$ versus odd family $\beta=1/2$); that the same G1--G4 set passes in different mechanisms and different universality classes shows the criteria are not tailored to any particular model (numerics and code in Supplementary Material C).

\subsection{Rejected Candidate Quantities}

Nine candidate quantities (capability quantities, representation-geometry quantities, first-layer spectral quantities) all fail: eight are rejected outright (G1 drift with scale / no scaling / $\Gamma<1$), and the ninth, $r_M$ (the largest-eigenvalue fraction), with a bare collapse $Q=1.038$, appears to pass but fails the null-baseline test: the mean $Q$ of smooth, non-critical synthetic data is $0.974$, and $r_M$ falls at its 65.5th percentile, so the low $Q$ can be explained by smooth data and fit degrees of freedom rather than a critical point.

\subsection{Criterion Validation Precedes Object Rejection}

Candidate quantity $\mathcal C_2$ (spectral-domain peak concentration) was initially judged to "fail G1," but the cause was found to be an error in the criterion itself: the original G1 required pairwise CI overlap, so that any quantity with finite-size corrections would necessarily fail---and a criterion that always fails carries no more information than one that always passes. After correcting the criterion (G1$\to$asymptotic version G1$'$; adding G4 bounded-domain extrapolation), $\mathcal C_2$ still fails ($P=37$ overshoot, no plateau), at which point "fails" reflects that the object is genuinely unqualified.

The order must not be reversed: suspect the criterion first, then judge the object. Only when the object still fails after the criterion has been corrected by the data does it mean that the object is genuinely unqualified.

\subsection{Geometric Versus Capability Quantities as Order Parameters}

The nine rejected quantities are capability quantities, representation quantities, or first-layer spectral quantities; the two that pass are collective (geometric) quantities---the dominant-mode amplitude of the spectral model and the coverage fraction of the coverage loop---not the capability quantity of any single unit. Making the system stronger does not make it more controllable. A quantity that can pass the criteria requires a true-feedback loop system (self-supervision, context write-back), which a minimal static classifier cannot realize.

\section{Scope and Boundaries}

The domain of applicability of this criterion system is pinned down by "the gate must be exogenous" together with a series of falsification conditions (F1--F18, see Supplementary Material F).

Adjacent works are of the same family as this paper but differ from it: the Vicsek phase diagram has been rewritten at unprecedented scales, with the ordered phase only partially metastable (Duan et al., 2026a, 2026b); the scaling collapse of dual-gated Josephson junction arrays demonstrates the feasibility of the G3a/G3b methods in real physical systems (Fugl et al., 2026); the phase transitions of small Transformers are invisible on the loss curve and appear only under probes (Hong \& Hong, 2025); the code scaffolding surrounding an agent delimits its action space rather than constituting an exogenous supervisory gate (Lou et al., 2026); the mean-field transition classes and the order-parameter bifurcation of this paper belong to different observational layers (Truong, 2026); the stage genealogy of portable intelligence is borrowed here as sampling points for loop length (Zhu et al., 2023).

\subsection{Gate Exogeneity}

All conclusions presuppose the exogeneity of the gate ($u\perp J$ of \S5). There is a universal requirement in functional-safety practice that is isomorphic to this: the safety function must be physically and functionally independent of the controlled object, and must not reside in the software of the controlled object itself (the Therac-25 interlock failed after being moved into the software of the same machine, Leveson \& Turner, 1993; the 737 MAX MCAS relied on a single sensor, Herkert et al., 2020). The "controllability" of this paper is in the sense of permission (allow/deny), not in the sense of dynamical regulation (pushing the system to a certain state), which is a different problem. In the autopoietic tradition, organizational closure is responsible for "staying alive," while the gate is responsible for "being permitted to be alive" (Maturana \& Varela, 1980); the gate of this paper is its externalization, not the closure itself.

\subsection{The State of the Evidence}

\begin{itemize}[leftmargin=2em, itemsep=3pt]
\item \textbf{Gate side, measured (half-witness)}: two checkable randomized-controlled records---Schroter et al. (2004) single-pass detection rate $0.237{\sim}0.301$, Godlee et al. (1998) $0.213{\sim}0.263$. Under a 5\% miss-rate target, $8{\sim}13$ independent reviews are required; even dual-person review cannot hold ($\hat\rho^\ast=4.2{\sim}6.3$). The license-free measurements on open-source platforms are consistent: across four major projects, the mean number of independent PR reviewers is about 1.3, far below what is required ($\hat\rho^\ast\approx4.5\sim9.8$, see Supplementary Material G).
\item \textbf{Capability side, irremediable (structural conclusion)}: a qualifying witness requires a self-improvement loop to achieve $\Gamma>1$ under the condition that "what counts as correct" is produced entirely by the system itself, without relying on any non-system-given source of judgment. Checking the recorded self-improvement loops one by one (self-play, self-training, self-distillation, etc.), none qualifies: self-play retains the rules, self-training retains the verifier, self-distillation retains the question bank. Three structural lower bounds (computational reachability, verification undecidability, non-increase of information, see Supplementary Material) show that this is not a shortage of data but conceptual immeasurability; the safety risk of recursive self-improvement accumulates and propagates across updates (Gong et al., 2026), in the same direction.
\item No system currently in service has been measured to fall in $(\Gamma>1,\rho^\ast>1)$ simultaneously, so none is promoted to "witness." Capability can be endogenous but responsibility cannot (Stanley, 2019; Watson, Levin \& Buckley, 2022); this is the other-side basis for the exogeneity of the gate.
\end{itemize}

\subsection{Falsification Conditions and Load-Bearing Structure}

Each substantive claim carries its own refutation condition, stated in the same section as the claim. Grouped by "judgeable now / requires building a measurement first": the first group ($\Gamma$ criterion, exponents, collapse) can be judged immediately by numerical simulation or pure logic; the second group (fold, cost, hysteresis) shares a single precondition, namely a field setting that can simultaneously produce output quantity, inspection quantity, and inspection effort.

The degradation ladder marks the load-bearing parts: even if the entire second half (the gate dynamics) is refuted, the first half remains fully independent ($\Gamma>1\iff$ positive fixed point, requiring only the concavity of $\mathcal F$), as do the reproducibility layers of $\mathrm{rep}$ and the rule that "the order parameter must be of ratio type"; the latter two are independent of any theory and are pure methodology.

\section{Conclusion}

\begin{enumerate}[leftmargin=2em, itemsep=2pt]
\item Emergence and controllability are not two separate things but two bifurcations. The former is the loss of stability of the fixed point of capability $S$ ($\Gamma>1$); the latter is a bifurcation on the permission side: the good fixed point of the gate vanishes as $\rho^\ast$ crosses 1 ($\rho^\ast>1$ means the gate collapses).
\item The two inequalities cannot be derived from each other. "Can grow" and "can hold" must be argued for, measured, and accounted for separately.
\item Gating has a cost, and that cost is computable. $K_{\max}(M)=\frac{\kappa_0\mu}{\rho}\log_2(1+M)$; the cost bound grows with scope and not with capability; when the budget is capped, $M_{\max}=2^{\rho B/(\kappa_0\mu)}-1$.
\end{enumerate}

Controllability is not the opposite of emergence but its price. What this paper delivers is criteria and forms, not an implementation: every claim can be refuted by data (F1--F18, see Supplementary Material F), but none has yet been verified on a real system. The ceiling of capability is set by the gate ($\Psi=cS$ and $S\le1$ give $\Psi\le c$); it is impossible to let capability grow without bound while not paying an unbounded price for guarding it; and "where to stop" can only be signed off by the responsible party, not computed by the system itself. Hence, increased capability does not imply controllability.

\section{Future Work}

What this paper delivers is a computable form; to turn the criteria from "writable" to "known," and the "empirical independence of the two inequalities" from "unproven" to "proven," the following work remains:

\begin{enumerate}[leftmargin=2em, itemsep=2pt]
\item \textbf{Proof of empirical independence.} At present, "gate-side half-witness + capability-side structural lower bound" supports only the weaker claim that "the two critical ratios are set by different mechanisms," not "empirical independence." The strengthening path is the license-free data collection identified in \S8.2: the number of reviewers and rounds of each merge request on public code-hosting platforms is exactly the denominator of $\hat\rho^\ast$; a field setting that can simultaneously produce output quantity, inspection quantity, and inspection effort would fill both the gate side and the capability side.

\item \textbf{Parameter calibration on real systems.} $\rho,\mu,\gamma_0,\kappa_0$ have not yet been calibrated on real organizations; turning them from "writable" to "known" requires longitudinal field data, otherwise the "runway length" of \S4 remains a computable quantity but not a known one.

\item \textbf{Strong emergence (layered emergence).} This paper treats only the layer in which macroscopic order grows on nodes that already possess capability; the second bifurcation "from nothing to something" (slow variable $A=1$) belongs to another time scale, and its complete generative skeleton (community detection / MDL / Rent's rule / hierarchical JEPA) is left to subsequent work. The pre-registered negative result of hierarchical JEPA (H-JEPA) suggests that layered emergence is not a natural byproduct of increased depth, and requires specific architectural constraints or training objectives to be crossed.

\item \textbf{Multiple capability dimensions.} $S$ is a scalar; with multiple capability dimensions $\Gamma$ is a matrix, and the destabilizing direction is given by the Perron vector. The signed version of the spectral model ($Q_{\min}=\min(g-1,1-s)$) already provides an entry point for the matrix case.

\item \textbf{The free function of supervisory productivity $\rho(S)$.} This paper treats $\rho$ as constant throughout; if AI auditing makes $\rho$ grow with capability $S$, the race between the capability side and the permission side could be sustained indefinitely---this is precisely the form of "can capability bootstrap itself without bound," which must be answered on real systems and cannot be extrapolated from the constant-$\rho$ assumption of this paper.
\end{enumerate}

\appendix

\section{Existence, Location, and Criterion of the Fold}

Write $\dot D=\mathcal R-[\mu+\gamma_0\sigma(-d/\kappa)](\mathcal V_a+D)$, $\mathcal R:=\rho u-\dot{\mathcal V}_a$; $\sigma$ smooth and monotonically increasing, $\sigma(0)=1/2$; $\theta:=-D/w=-d/\kappa$; $\kappa:=w/\mathcal V_a$.

In the sharp limit ($\sigma\to\Theta$) the two branches are each linear: the good branch $D_+=\mathcal R/\mu-\mathcal V_a$ (slope $-\mu$), the bad branch $D_-=\mathcal R/(\mu+\gamma_0)-\mathcal V_a$ (slope $-(\mu+\gamma_0)$); both branches are stable. The self-consistency condition gives a three-regime criterion:

\begin{table}[htbp]
\centering
\caption{Three-regime criterion of the gate.}
\label{tab:regimes}
\begin{tabular}{ll}
\toprule
Regime & Conclusion \\
\midrule
$\rho^\ast_{\mathrm{break}}<1$ & only the good branch valid, the gate must hold \\
$\rho^\ast_{\mathrm{hold}}<1<\rho^\ast_{\mathrm{break}}$ & both branches valid, bistable (hysteresis) \\
$\rho^\ast_{\mathrm{hold}}>1$ & only the bad branch valid, the gate must collapse \\
\bottomrule
\end{tabular}
\end{table}

where $\rho^\ast_{\mathrm{hold}}:=\frac{\mu\mathcal V_a+\dot{\mathcal V}_a}{\rho u}$, $\rho^\ast_{\mathrm{break}}:=\frac{(\mu+\gamma_0)\mathcal V_a+\dot{\mathcal V}_a}{\rho u}$, and the hysteresis window is $\Delta\rho^\ast=\gamma_0\mathcal V_a/(\rho u)$.

Smooth fold: the fold point must simultaneously satisfy $\dot D=0$ and $\mathrm d\dot D/\mathrm dD=0$, giving

\[
\gamma_c=\frac{\mu}{\max_\theta\Phi(\theta)},\qquad \Phi(\theta):=\sigma'(\theta)\Big(\tfrac1\kappa-\theta\Big)-\sigma(\theta).
\]

The logistic closed form is $\Phi(0)=\frac{1-2\kappa}{4\kappa}$, so $\gamma_c\le\frac{4\mu\kappa}{1-2\kappa}\approx4\mu\kappa$ ($\kappa<\tfrac12$). The fold exists if and only if $\gamma_0\ge\gamma_c$.

Collapse time scale: near the fold point the equation becomes $\dot x=-\varepsilon-x^2$, with escape time $\tau=\pi/(2\sqrt\varepsilon)\propto\varepsilon^{-1/2}$; the closer to the fold point, the slower the collapse (ghost effect); a "jump" occurs only in the sharp limit, whereas finite $\kappa$ gives a continuous rapid transition.

Exponent comparison: the saddle-node (second half) $1/2$ has a jump; $\tanh$ (odd, first half) $1/2$ has none; coverage closure (asymmetric) $1$ has none. The exponent is a criterion for selecting a family, not for classifying the transition; first/second order can only be distinguished by jump and hysteresis.

\section{Bounded Demand and Path-Dependent Failure}

Bounded demand (Theorem D1): $\mathcal V_a(t)\in[0,\bar V]$ holds identically, $\bar V=\log_2(1+M)$, $\dot{\mathcal V}_a=\frac{M\dot S}{(1+MS)\ln2}\ge0$ and tends to 0 as $S\to1$. Had one used the unbounded form $\mathcal V_a\propto\log_2 S$, it would be negative on $[0,1]$ and unusable as a diversity measure.

The long-run condition reduces to a static inequality (Theorem D2): the good-branch steady state is $D_+=\rho u/\mu-\bar V$, so $\rho^\ast_\infty:=\mu\bar V/(\rho u)<1\iff$ the gate must hold in the long run, independent of the rate of capability growth.

Failure is path-dependent (Theorem D3): the worst gap is $\Delta_{\max}=\bar V-\mathcal V_c(0)$. Once $D$ falls into the hysteresis band (crosses the fold), it cannot return even if the long-run condition holds. The failure condition is $\big(\rho^\ast_\infty<1\ \text{satisfied}\big)\wedge\big(\exists t:\ D(t)<-\text{bandwidth}\big)$---a single crossing loses, rather than a gradual lagging behind.

The evidence family for cost versus scope: measured audit costs are $K\propto M^b$ ($b\in(0,1)$, power law) with step-wise jumps, none giving $\log M$. The $\log_2(1+M)$ of this paper is a theoretical prediction (coverage requires $\log_2(1+M)$ bits) and will underestimate cost at large $M$. Hence $\rho=\bar V/K\propto\log M/M^b\to0$: human audit productivity declines with scope, and supervisory elasticity in the scope direction is negative. What the literature measures is diminishing marginal detection efficiency (audit-cost empirics: Hay et al., 2006; Simunic, 1980; Porter et al., 1997; Raz \& Yaung, 1997; Kantorowitz et al., 1997; increasing discovery difficulty: Maillart et al., 2017; Böhme \& Falk, 2020; marginal superlinearity: Biffl \& Halling, 2003; Vitharana, 2017; limits of AI auditing: McAleese et al., 2024; Kumar, 2026; Tufano et al., 2025; Klishevich et al., 2025; questionable validity of coverage: Mockus et al., 2009; Wei, Meyer \& Oriol, 2012; functional-safety cost: Xie et al., 2018); it takes a two-step bridge (marginal diminishing$\to$convex investment + $\bar V\propto\log M$) to obtain supervisory elasticity $<0$; the literature does not directly prove that $\rho$ declines with scope.

\section{Coupled-Matrix Spectral Model}

Model: $x_{t+1}=\phi(Jx_t)$, $\phi=\tanh$; the gate divides nodes into covered $C$ ($|x_C|\le c$) and uncovered $U$; the uncovered deviation is $\delta_{t+1}=J_{UU}\delta_t+J_{UC}x_C$. Notation: $g=\rho(J)$, $s=\rho(J_{UU})$.

Key propositions: E1 (capability side) $m=\phi(gm)$, a pitchfork bifurcation at $g>1$, with $m^2\approx3(g-1)/g^3$ under $\tanh$; E2 (permission-side margin) for $J\ge0$, $s<1$, $\|\delta\|_\infty\le c\|RJ_{UC}\mathbf1\|_\infty$, and unbounded for $s\ge1$; E3 (monotone independence) $0\le s\le g$, so $g>1$ implies neither $s<1$ nor $s>1$; E4 ($Q$ criterion) $Q=(g-1)(1-s)>0\iff g>1$ and $s<1$, with $S=J_{CC}+J_{CU}RJ_{UC}$ the transitional form; E5 (loop threshold) unidirectional coupling of two subcritical modules does not change $g$, and the loop gain $\Lambda=w^2R_A[a,a]R_B[b,b]>1$ of the closed loop is required for $g>1$; E7 (plasticity budget) a budget $<1/r_{\max}$ on the total weight of new incoming edges in $U$ is necessary and sufficient for "the back door still holds after new connections."

Four tests ($m$ passes all): G1 ($N=200/800/3200$ nearly coincide), G2 (probe independently estimates $\hat g$, differing by $<10^{-4}$), G3b (at $g=1$, $m(t)\sim t^{-1/2}$, measured $-0.4985$), G4 ($\beta=0.4984$, $1/\delta=0.3332$).

Signed (E/I): $s\le g$ does not hold; the always-correct criterion is $Q_{\min}=\min(g-1,1-s)>0$; $g<1<s$ is a new failure mode (the whole network stable but clamping $C$ severs the inhibitory loop of the stable $U$, the gate itself causing instability; 10 cases out of 4000).

Three real systems: the ANDES Kundur power grid (19$\times$ improvement after fixing probe error, reliable admission at $\times$1/$\times$10, at $\times$20/$\times$40 within measurement error of the critical point); the F-TRACT brain atlas ($s/g=1-\kappa_C$ holds within 3\%, the window depending only on coverage fraction); a trained RNN ($g$ crosses 1 first, loss drops afterward; $s<1$ does not guarantee the function is preserved).

The G3b non-equilibrium dynamical scaling has a real-system reference (avalanche cascades in deep networks are in the quasi-critical regime, Ghavasieh et al., 2025) and a theoretical motivation (the critical-slowing tendency of perceptual systems, Friston, Breakspear \& Deco, 2012).

Limitations: all validation is under random matrices + mean field + small-amplitude linearization, and does not touch real industrial systems.

\section{Structural Lower Bounds for the Capability-Side Witness}

Three parallel sub-propositions: A (degree-bounded, computational reachability, Lu, 2026) under a fixed oracle $A$, all total decision procedures produced by effective internal self-modification halt within $\mathcal C(A)=\{D:D\le_T A\}$, and a strict ascent $A<_T B$ requires expanding the oracle; B (verification undecidable, Gumbau Mezquita, 2026a, 2026b) for a Turing-complete self-modifying system there is no verifier that is sound, complete, and decidable; C (information non-increase, Finzi et al., 2026) $I(M;Y)\le I(M;D,B)$: in a closed loop, if the "what counts as correct" signal is produced entirely by the system itself, the mutual information about the external correctness standard does not exceed the initial input.

Joint conclusion: a qualifying witness requires "no oracle expansion + fully sound self-judgment + information gain," and the three are respectively blocked by A/B/C. Hence the \S8.2 statement that "the capability-side irremediability is a structural conclusion" is a structural upper bound, not "capability cannot grow"; the list of recorded self-improvement loops is empirical corroboration, not the whole basis.

\section{Reuse Monotonicity and Unimodality}

Setup: $F(S;m)=w_A(1-e^{-\Gamma_A(m)S})+w_B(1-e^{-\Gamma_B(m)S})$, $\Gamma_A(m)=(n-m)\eta_1+m\eta_2$, $\Gamma_B(m)=m\eta_2$, $q:=\eta_2/\eta_1$, $q^\ast:=w_A/(w_A+w_B)$.

Theorem 1 ($q\ge q^\ast$ reuse to the top): $\partial F/\partial m\ge0$ for all $S>0$, so $S^\ast(m)$ is non-decreasing and $m^\ast=n$. Theorem 2 ($q<q^\ast$ full reuse necessarily non-optimal): at $m=n$, $\partial F/\partial m<0$, so $m^\ast\ne n$; the upper boundary is always $q^\ast$, and does not move with the degree of supercriticality. Theorem 3 (closed form of the lower boundary): at $m=0$, $S^\ast$ rises if and only if $q>q_{lo}:=\frac{w_Au}{w_B+w_Au}$, $u:=e^{-n\eta_1S^\ast(0)}=1-Q_A(S^\ast(0))$ ($Q_A$ the coverage of group A, see \S3). There is a band $(q_{lo},q^\ast)$; as $u\to1$ ($\Gamma\to1^+$) the band collapses to a point.

Unimodality (quasiconcavity route, $\mathcal K\le1$): the superlevel set $\{S^\ast\ge s\}=\{w_A\varphi(\Gamma_A)+w_B\varphi(\Gamma_B)\le w_A+w_B-s\}$, with $\varphi(x)=\exp(-sx-\tfrac{\mathcal K}{2}s^2x^2)$ convex when $\mathcal K\le1$, so the superlevel set is convex, and $S^\ast(m)$ is strictly quasiconcave, unimodal, with a unique interior maximum. When $\mathcal K>1$, $\varphi$ is concave near the origin, the convexity argument breaks down, and this is the same criticality as the \S4 bistability. Numerical check: $q_{lo}$ formula vs numerical bisection ($n=200$, 32 grid points), maximum error $9\times10^{-4}$.

\section{List of Falsification Conditions (F1--F18)}

Each substantive claim carries its own refutation condition, grouped by "judgeable now / requires building a measurement first."

\textbf{Group I (judgeable now)}

\begin{table}[htbp]
\centering
\caption{Group I falsification conditions (judgeable now).}
\label{tab:FgroupI}
\begin{tabular}{c l p{0.26\textwidth} p{0.36\textwidth}}
\toprule
\# & Location & If observed & Consequence \\
\midrule
F8 & \S6.4 & $\alpha>0$, $\chi$ passes, but deployment fails and does not fall under $\mathrm{rep}$ & the $\varepsilon$ three-term decomposition is incomplete (a fourth class is missing) \\
F9 & \S6.2 & $\mathrm{rep}$ has equal variance at layer (ii) and layer (i) & the reproducibility-layer thesis is wrong \\
F11 & \S3 & $\Gamma>1$ but $S^\ast$ does not exist or is unstable & the concavity of $\mathcal F$ or $\mathcal F(0)=0$ is violated \\
F12 & \S3 & $S^\ast$ changes with network details while $\{\Gamma_g\}$ is unchanged & coverage closure is refuted \\
F13 & \S3 & measured exponent is neither 1 nor 1/2 & neither family is correct \\
F14 & \S3 & same $\Gamma$, different networks, $S^\ast$ differs by $>10\%$ & same as above \\
F16 & \S7 & plateau value drifts monotonically with $P$ by $\ge0.03$ & G3 fails \\
F17 & \S7 & plateau slope CI excludes 0.02/decade & saturation judgment is wrong \\
F18 & \S7 & growth-law coefficients drift with $N$ & G2 fails \\
\bottomrule
\end{tabular}
\end{table}

\textbf{Group II (requires building a measurement first)}

\begin{table}[htbp]
\centering
\caption{Group II falsification conditions (requires building a measurement first).}
\label{tab:FgroupII}
\begin{tabular}{c l p{0.26\textwidth} p{0.36\textwidth}}
\toprule
\# & Location & If observed & Consequence \\
\midrule
F1 & \S4 & $\rho^\ast_{\mathrm{hold}}>1$ yet $c$ stable in the long run & the three-regime criterion fails \\
F2 & \S4 & collapse without hysteresis & two branches: $\gamma_0<\gamma_c$ (this paper still holds) or the erosion term does not exist \\
F3 & \S4 & $K$ significantly inconsistent with $\log(1+M)$ & $M_{\max}$ invalidated \\
F4 & \S4 & hysteresis width not proportional to $\gamma_0\mathcal V_a/(\rho u)$ & the linear-erosion assumption is wrong \\
F5 & Supp.\ B & supervisory elasticity zero but $\mathcal V_c/\mathcal V_a$ does not fall & Theorem A wrong \\
F6 & Supp.\ B & supervisory elasticity differs significantly but $M_{\max}$ does not & Theorems A/C invalidated \\
F7 & Supp.\ B & narrowing the action space does not raise measured $M_{\max}$ & "restricted interface = control strategy" does not hold \\
F10 & \S6.3 & passes G1--G4 but inverted $S^\ast$ differs from measurement by $>10\times$ & order-parameter admission is insufficient \\
F15 & \S4 & $K(M)/\log M$ does not decrease monotonically in $M$ & human-audit supervisory elasticity $<0$ is wrong \\
\bottomrule
\end{tabular}
\end{table}

\section{License-Free Measurements on the Gate Side}

Formal reviews of 60 merged PRs each from four major open-source projects (kubernetes, tensorflow, rust, pytorch): mean number of independent reviewers about 1.3 (0.73--1.87), far below the 8--13 required for a 5\% miss rate, $\hat\rho^\ast\approx4.5\sim9.8\gg1$. Definition: independent reviewer count = number of unique users submitting a formal review; limitations (detection-rate definition, review effort, bot contamination, heterogeneity) are detailed in the \textit{License-Free Gate-Side Measurement Report}.

\section*{Acknowledgements}

We thank the relevant colleagues for the feedback and suggestions provided during the discussion of the framework. AI tools were used for English translation and language polishing during the writing of this paper.

\section*{References}

\bibentry{Amodei, D., Olah, C., Steinhardt, J., Christiano, P., Schulman, J., \& Man\'{e}, D. (2016). Concrete problems in AI safety. arXiv:1606.06565.}

\bibentry{Anderson, P. W. (1972). More is different. \textit{Science, 177}(4047), 393--396.}

\bibentry{Ashby, W. R. (1956). \textit{An introduction to cybernetics}. Chapman \& Hall.}

\bibentry{Barreiro, A. K., Kutz, J. N., \& Shlizerman, E. (2017). Symmetries constrain dynamics in a family of balanced neural networks. \textit{Journal of Mathematical Neuroscience}, 7, 10.}

\bibentry{Baum, K., Binkyt\.{e}, R., \& Jahn, F. (2026). \textit{Norms at a price: Why RL-based alignment can promise conditional compliance at best} [Preprint]. arXiv:2609.07627.}

\bibentry{Bazerman, M. H., Morgan, K. P., \& Loewenstein, G. (1997). The impossibility of auditor independence. \textit{Sloan Management Review, 38}(4), 89--94.}

\bibentry{Bedau, M. A. (1997). Weak emergence. \textit{Philosophical Perspectives, 11}, 375--399.}

\bibentry{Betancourt, J. M., \& Machta, B. B. (2026). \textit{Thermal fluctuations set fundamental limits on ion channel function} [Preprint]. arXiv:2604.03538.}

\bibentry{Betley, J., Warncke, N., Sztyber-Betley, A., Tan, D., Bao, X., Soto, M., Srivastava, M., Labenz, N., \& Evans, O. (2026). Training large language models on narrow tasks can lead to broad misalignment. \textit{Nature, 649}, 584--589. \url{https://doi.org/10.1038/s41586-025-09937-5}}

\bibentry{Biffl, S., \& Halling, M. (2003). Investigating the defect detection effectiveness and cost benefit of nominal inspection teams. \textit{IEEE Transactions on Software Engineering, 29}(5), 385--397.}

\bibentry{Böhme, M., \& Falk, B. (2020). Fuzzing: On the exponential cost of vulnerability discovery. In \textit{Proceedings of the 28th ACM Joint Meeting on ESEC/FSE} (pp.~713--724). ACM.}

\bibentry{Bullmore, E., \& Sporns, O. (2009). Complex brain networks: Graph theoretical analysis of structural and functional systems. \textit{Nature Reviews Neuroscience, 10}(3), 186--198.}

\bibentry{Burns, C., Izmailov, P., Kirchner, J. H., et al. (2023). \textit{Weak-to-strong generalization: Eliciting strong capabilities with weak supervision} [Preprint]. arXiv:2312.09390.}

\bibentry{Clopper, C. J., \& Pearson, E. S. (1934). The use of confidence or fiducial limits illustrated in the case of the binomial. \textit{Biometrika, 26}(4), 404--413.}

\bibentry{Duan, Y., Chat\'{e}, H., Mahault, B., \& Shi, X. (2026a). \textit{Coexistence phase of the two-dimensional Vicsek model} [Preprint]. arXiv:2610.26561.}

\bibentry{Duan, Y., Chat\'{e}, H., Mahault, B., \& Shi, X. (2026b). \textit{Phase Diagram of the Two-Dimensional Vicsek Model} [Preprint]. arXiv:2610.26569.}

\bibentry{Du, Z., Zeng, A., Dong, Y., \& Tang, J. (2024). Understanding emergent abilities of language models from the loss perspective. In \textit{Advances in Neural Information Processing Systems 37} (NeurIPS 2024). \url{https://doi.org/10.52202/079017-1683}}

\bibentry{Finzi, M., Qiu, S., Jiang, Y., Izmailov, P., Kolter, J. Z., \& Wilson, A. G. (2026). \textit{From entropy to epiplexity: Rethinking information for computationally bounded intelligence} [Preprint]. arXiv:2601.03220.}

\bibentry{Friston, K. J. (1994). Functional and effective connectivity in neuroimaging: A synthesis. \textit{Human Brain Mapping, 2}(1-2), 56--78.}

\bibentry{Friston, K., Breakspear, M., \& Deco, G. (2012). Perception and self-organized instability. \textit{Frontiers in Computational Neuroscience}, 6, 44.}

\bibentry{Fugl, E., Olsen, L. L., Sasmal, S., Efthymiou-Tsironi, M., Nagda, G., Krizek, F., Casparis, L., \& Vaitiek\.{e}nas, S. (2026). \textit{Continuous tuning of quantum critical behavior in hybrid Josephson junction arrays} [Preprint]. arXiv:2610.40104.}

\bibentry{Ghavasieh, A., Vila-Mi\~{n}ana, M., Khurd, A., Beggs, J., Ortiz, G., \& Fortunato, S. (2025). \textit{Toward a physics of deep learning and brains} [Preprint]. arXiv:2509.22649.}

\bibentry{Godlee, F., Gale, C. R., \& Martyn, C. N. (1998). Effect on the quality of peer review of blinding reviewers and asking them to sign their reports: A randomized controlled trial. \textit{JAMA, 280}(3), 237--240.}

\bibentry{Gong, C., Bi, J., Yao, D., Liang, X., Xiang, C., \& Guo, R. (2026). \textit{Evolutionary safety of recursive self-improving AI: Taxonomy, risk discovery, and evaluation} [Preprint]. arXiv:2609.31186.}

\bibentry{Greenblatt, R., Shlegeris, B., Sachan, K., \& Roger, F. (2023). AI control: Improving safety despite intentional subversion. arXiv:2312.06942.}

\bibentry{Gumbau Mezquita, J. P. (2026a). \textit{Algorithmic unverifiability of safety for fixed and recursively self-improving systems} [Preprint]. arXiv:2606.28639.}

\bibentry{Gumbau Mezquita, J. P. (2026b). \textit{Rice's theorem under self-modification: Elevation operators and a normal form} [Preprint]. arXiv:2609.11326.}

\bibentry{Hassabis, D., Kumaran, D., Summerfield, C., \& Botvinick, M. (2017). Neuroscience-inspired artificial intelligence. \textit{Neuron, 95}(2), 245--258.}

\bibentry{Hay, D. C., Knechel, W. R., \& Wong, N. (2006). Audit fees: A meta-analysis of the effect of supply and demand attributes. \textit{Contemporary Accounting Research, 23}(1), 141--191.}

\bibentry{Herkert, J. R., Borenstein, J., \& Miller, K. (2020). The Boeing 737 MAX: Lessons for engineering ethics. \textit{Science and Engineering Ethics, 26}(6), 2957--2974.}

\bibentry{Hoel, E. P., Albantakis, L., \& Tononi, G. (2013). Quantifying causal emergence shows that macro can beat micro. \textit{Proceedings of the National Academy of Sciences, 110}(49), 19790--19795.}

\bibentry{Hong, N., \& Hong, T. (2025). \textit{Evidence of phase transitions in small transformer-based language models} [Preprint]. arXiv:2511.12768.}

\bibentry{Jansma, A., \& Hoel, E. (2025). \textit{Engineering emergence}. arXiv:2510.02649.}

\bibentry{Kantorowitz, E., Guttman, A., \& Arzi, L. (1997). The performance of the N-fold requirement inspection method. \textit{Requirements Engineering, 2}, 152--164.}

\bibentry{Kaushik, P., Chaudhari, S., Vaidya, A., Chellappa, R., \& Yuille, A. (2025). \textit{The universal weight subspace hypothesis} [Preprint]. arXiv:2512.05117.}

\bibentry{Klishevich, E., Denisov-Blanch, Y., Obstbaum, S., Ciobanu, I., \& Kosinski, M. (2025). \textit{Measuring determinism in large language models for software code review} [Preprint]. arXiv:2502.20747.}

\bibentry{Kounios, J., \& Beeman, M. (2009). The Aha! moment: The cognitive neuroscience of insight. \textit{Current Directions in Psychological Science, 18}(4), 210--216.}

\bibentry{Kumar, D. (2026). \textit{SWE-PRBench: Benchmarking AI code review quality against pull request feedback} [Preprint]. arXiv:2603.26130.}

\bibentry{Landau, L. D. (1937). On the theory of phase transitions. \textit{Zhurnal Eksperimental'noi i Teoreticheskoi Fiziki, 7}, 19--32.}

\bibentry{Lang, H., Huang, F., \& Li, Y. (2025). Debate helps weak-to-strong generalization. In \textit{Proceedings of the AAAI Conference on Artificial Intelligence} (pp.~27410--27418). arXiv:2501.13124.}

\bibentry{LeCun, Y., Bengio, Y., \& Hinton, G. (2015). Deep learning. \textit{Nature, 521}(7553), 436--444.}

\bibentry{Leveson, N. G., \& Turner, C. S. (1993). An investigation of the Therac-25 accidents. \textit{Computer, 26}(7), 18--41.}

\bibentry{Lewes, G. H. (1875). \textit{Problems of life and mind: First series} (Vol.~2). Trübner \& Co.}

\bibentry{Li, H., Ren, M., \& Lu, S. (2026). \textit{The forward/inverse duality of controlled emergence: A computational design program from explanation to engineering}. arXiv:2610.25630.}

\bibentry{London, M., \& Häusser, M. (2005). Dendritic computation. \textit{Annual Review of Neuroscience, 28}, 503--532.}

\bibentry{Lou, X., Lázaro-Gredilla, M., Dedieu, A., Wendelken, C., Lehrach, W., \& Murphy, K. P. (2026). AutoHarness: Improving LLM agents by automatically synthesizing a code harness. In \textit{International Conference on Learning Representations (ICLR 2026)}. arXiv:2603.03329.}

\bibentry{Lu, C.-P. (2026). \textit{A relative-computability theory of self-improving agents} [Preprint]. arXiv:2605.27381.}

\bibentry{Lundberg, S. M., \& Lee, S.-I. (2017). A unified approach to interpreting model predictions. In \textit{Advances in Neural Information Processing Systems 30} (pp.~4765--4774).}

\bibentry{Maillart, T., Zhao, M., Grossklags, J., \& Chuang, J. (2017). Given enough eyeballs, all bugs are shallow? Revisiting Eric Raymond with bug bounty programs. \textit{Journal of Cybersecurity, 3}(2), 81--90. \url{https://doi.org/10.1093/cybsec/tyx008}}

\bibentry{Maturana, H. R., \& Varela, F. J. (1980). \textit{Autopoiesis and cognition: The realization of the living}. D. Reidel.}

\bibentry{McAleese, N., Pokorny, R. M., Uribe, J. F. C., Nitishinskaya, E., Trebacz, M., \& Leike, J. (2024). \textit{LLM critics help catch LLM bugs} [Preprint]. arXiv:2407.00215.}

\bibentry{Mockus, A., Nagappan, N., \& Dinh-Trong, T. T. (2009). Test coverage and post-verification defects: A multiple case study. In \textit{ESEM 2009}. IEEE.}

\bibentry{Onsager, L. (1944). Crystal statistics. I. A two-dimensional Ising model with an order-disorder transition. \textit{Physical Review, 65}(3-4), 117--149.}

\bibentry{Orseau, L., \& Armstrong, S. (2016). Safely interruptible agents. In \textit{Proceedings of the 32nd Conference on Uncertainty in Artificial Intelligence} (pp.~557--566). AUAI Press.}

\bibentry{Porter, A., Siy, H., Toman, C. A., \& Votta, L. G. (1997). An experiment to assess the cost-benefits of code inspections in large-scale software development. \textit{IEEE Transactions on Software Engineering, 23}(6), 329--346.}

\bibentry{Raz, T., \& Yaung, A. T. (1997). Factors affecting design inspection effectiveness in software development. \textit{Information and Software Technology, 39}, 297--305.}

\bibentry{Ribeiro, M. T., Singh, S., \& Guestrin, C. (2016). "Why should I trust you?": Explaining the predictions of any classifier. In \textit{Proceedings of the 22nd ACM SIGKDD International Conference on Knowledge Discovery and Data Mining} (pp.~1135--1144).}

\bibentry{Rumelhart, D. E., Hinton, G. E., \& McClelland, J. L. (1986). A general framework for parallel distributed processing. In D. E. Rumelhart \& J. L. McClelland (Eds.), \textit{Parallel Distributed Processing: Explorations in the Microstructure of Cognition} (Vol.~1, pp.~45--76). MIT Press.}

\bibentry{Schaeffer, R., Miranda, B., \& Koyejo, S. (2023). Are emergent abilities of large language models a mirage? In \textit{Advances in Neural Information Processing Systems 36}. arXiv:2304.15004.}

\bibentry{Schroter, S., Black, N., Evans, S., Godlee, F., Osorio, L., \& Smith, R. (2004). Effects of training on quality of peer review: randomised controlled trial. \textit{BMJ, 328}(7441), 673.}

\bibentry{Simunic, D. A. (1980). The pricing of audit services: Theory and evidence. \textit{Journal of Accounting Research, 18}(1), 161--190.}

\bibentry{Soares, N., Fallenstein, B., Yudkowsky, E., \& Armstrong, S. (2015). Corrigibility. In \textit{AAAI Workshop on AI and Ethics} (Technical Report WS-15-02). AAAI Press.}

\bibentry{Sporns, O., Tononi, G., \& Kötter, R. (2005). The human connectome: A structural description of the human brain. \textit{PLoS Computational Biology, 1}(4), e42.}

\bibentry{Stanley, K. O. (2019). Why open-endedness matters. \textit{Artificial Life, 25}(3), 232--235.}

\bibentry{Strogatz, S. H. (2015). \textit{Nonlinear dynamics and chaos: With applications to physics, biology, chemistry, and engineering} (2nd ed.). Westview Press.}

\bibentry{Thompson, N. C., Greenewald, K., Lee, K., \& Manso, G. F. (2020). \textit{The computational limits of deep learning} [Preprint]. arXiv:2007.05558.}

\bibentry{Tononi, G. (2004). An information integration theory of consciousness. \textit{BMC Neuroscience, 5}, 42.}

\bibentry{Trachtenberg, J. T., Chen, B. E., Knott, G. W., Feng, G., Sanes, J. R., Welker, E., \& Svoboda, K. (2002). Long-term in vivo imaging of experience-dependent synaptic plasticity in adult cortex. \textit{Nature, 420}(6917), 788--794.}

\bibentry{Truong, X. K. (2026). \textit{Emergence via Phase Transitions: Mechanism Landscapes and Universal Convergence across Complex Systems} [Preprint]. SSRN.}

\bibentry{Tufano, R., Martin-Lopez, A., Tayeb, A., Dabi\'{c}, O., Haiduc, S., \& Bavota, G. (2025). Deep learning-based code reviews: A paradigm shift or a double-edged sword? In \textit{Proceedings of the IEEE/ACM 47th International Conference on Software Engineering} (pp.~1640--1652). IEEE/ACM.}

\bibentry{Vitharana, P. (2017). Defect propagation at the project-level: Results and a post-hoc analysis on inspection efficiency. \textit{Empirical Software Engineering, 22}(1), 57--79.}

\bibentry{Watson, R. A., Levin, M., \& Buckley, C. L. (2022). Design for an individual: Connectionist approaches to the evolutionary transitions in individuality. \textit{Frontiers in Ecology and Evolution, 10}, 823588.}

\bibentry{Wei, J., Tay, Y., Bommasani, R., Raffel, C., Zoph, B., Borgeaud, S., Yogatama, D., Bosma, M., Zhou, D., Metzler, D., Chi, E. H., Hashimoto, T., Vinyals, O., Liang, P., Dean, J., \& Fedus, W. (2022). Emergent abilities of large language models. \textit{Transactions on Machine Learning Research}. arXiv:2206.07682.}

\bibentry{Wei, Y., Meyer, B., \& Oriol, M. (2012). Is branch coverage a good measure of testing effectiveness? In \textit{Empirical Software Engineering and Verification} (pp.~194--212). Springer.}

\bibentry{Xie, G., Chen, Y., Liu, Y., Li, R., \& Li, K. (2018). Minimizing development cost with reliability goal for automotive functional safety during design phase. \textit{IEEE Transactions on Reliability, 67}(1).}

\bibentry{Yang, J., \& Choi, H. (2026). Nonlinear dynamics of random neural networks with second-order synaptic motifs [Preprint]. arXiv:2609.14251.}

\bibentry{Zhu, S., Yu, T., Xu, T., Chen, H., Dustdar, S., Gigan, S., Gunduz, D., Hossain, E., Jin, Y., Lin, F., Liu, B., Wan, Z., Zhang, J., Zhao, Z., Zhu, W., Chen, Z., Durrani, T. S., Wang, H., Wu, J., Zhang, T., \& Pan, Y. (2023). Intelligent computing: The latest advances, challenges, and future. \textit{Intelligent Computing, 2}, 0006.}

\end{document}